\documentclass[10pt,twocolumn,letterpaper]{article}

\usepackage[pagenumbers]{cvpr} %

\usepackage{booktabs}
\usepackage{graphicx}
\usepackage{subcaption}
\usepackage{wrapfig}
\usepackage{colortbl}
\usepackage{comment}
\usepackage{adjustbox}

\newcommand{\methodtitleshort}{\textit{\hbox{GenZoo}}\xspace}
\newcommand{\testdatasettitle}{\textit{\hbox{GenZoo-Felidae}}\xspace}

\usepackage{dblfloatfix}

\newcommand*{\affaddr}[1]{#1} 
\newcommand*{\affmark}[1][*]{\textsuperscript{#1}}
\newcommand*{\email}[1]{\small{\texttt{#1}}}

\definecolor{cvprblue}{rgb}{0.21,0.49,0.74}
\usepackage[pagebackref,breaklinks,colorlinks,allcolors=cvprblue]{hyperref}

\def\paperID{1398} %
\def\confName{CVPR}
\def\confYear{2025}

\title{Generative Zoo}

\author{%
Tomasz~Niewiadomski\affmark[1] \quad
Anastasios~Yiannakidis\affmark[1] \quad
Hanz~Cuevas-Velasquez\affmark[1] \quad
Soubhik~Sanyal\affmark[1]\\
Michael~J.~Black\affmark[1] \quad
Silvia~Zuffi\affmark[2] \quad
Peter~Kulits\affmark[1]\\
{\small \affaddr{\affmark[1]Max Planck Institute for Intelligent Systems, T{\"u}bingen, Germany} \quad \affaddr{\affmark[2]IMATI-CNR, Milan, Italy}}\\
\tt\small{
\{tomasz,ayiannakidis,hcuevas,ssanyal,black,kulits\}@tue.mpg.de,\quad \email{silvia.zuffi@cnr.it}
\vspace{-2.0em}
}
}

\begin{document}
\newcommand{\teaserCaption}{
We propose a pipeline for the scalable generation of realistic 3D animal pose and shape estimation training data.
Training solely on data produced using our pipeline, we achieve state-of-the-art performance on a real-world 3D pose and shape estimation benchmark.

}

\twocolumn[{
    \renewcommand\twocolumn[1][]{#1}
    \maketitle
    \centering
    \begin{minipage}{1.00\textwidth}
        \centering
        \includegraphics[width=\linewidth]{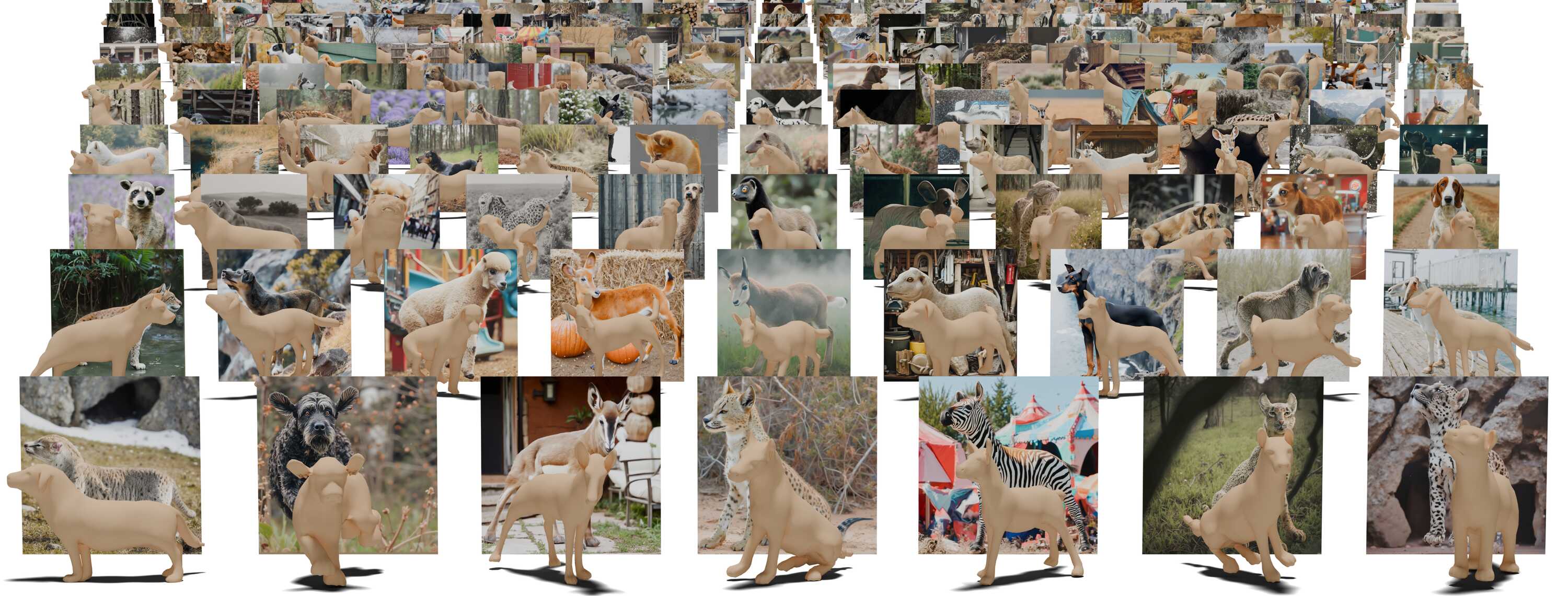}
    \end{minipage}
    \captionof{figure}{\teaserCaption}
    \label{fig:teaser}
    \vspace{0.5cm}
}]

\begin{abstract}
The model-based estimation of 3D animal pose and shape from images enables computational modeling of animal behavior.
Training models for this purpose requires large amounts of labeled image data with precise pose and shape annotations.
However, capturing such data requires the use of multi-view or marker-based motion-capture systems, which are impractical to adapt to wild animals in situ and impossible to scale across a comprehensive set of animal species.
Some have attempted to address the challenge of procuring training data by pseudo-labeling individual real-world images through manual 2D annotation, followed by 3D-parameter optimization to those labels.
While this approach may produce silhouette-aligned samples, the obtained pose and shape parameters are often implausible due to the ill-posed nature of the monocular fitting problem.
Sidestepping real-world ambiguity, others have designed complex synthetic-data-generation pipelines leveraging video-game engines and collections of artist-designed 3D assets.
Such engines yield perfect ground-truth annotations but are often lacking in visual realism and require considerable manual effort to adapt to new species or environments.
Motivated by these shortcomings, we propose an alternative approach to synthetic-data generation: rendering with a conditional image-generation model.
We introduce a pipeline that samples a diverse set of poses and shapes for a variety of mammalian quadrupeds and generates realistic images with corresponding ground-truth pose and shape parameters.
To demonstrate the scalability of our approach, we introduce \methodtitleshort, a synthetic dataset containing one million images of distinct subjects.
We train a 3D pose and shape regressor on \methodtitleshort, which achieves state-of-the-art performance on a real-world 3D animal pose and shape estimation benchmark, despite being trained solely on synthetic data.
We will make public our dataset and data-generation pipeline to support future \hbox{research at \url{https://genzoo.is.tue.mpg.de}.}
\end{abstract}

\section{Introduction}\label{sec:introduction}
The estimation of animal pose from images enables the computational modeling of animal behavior~\cite{anderson_toward_2014}.
In quantifying behavior, pose offers a low-dimensional representation amenable to analysis~\cite{pereira_quantifying_2020}.
From pose, actions can be segmented~\cite{MARSHALL2022102522} or individual health can be monitored~\cite{broome_going_2023}.

Pose alone is a fairly descriptive feature when measured in a laboratory setting, where cameras, lighting, and environmental conditions can be tightly controlled.
However, pose becomes less informative in the wild, where environmental conditions can vary greatly.
Inspired by trends in the modeling of humans~\cite{loper2015smpl,li_flame}, recent work has expanded beyond primitive pose-based representations and toward parametric models that represent not only 3D pose, but also shape~\cite{Zuffi_2017_CVPR,badger_2020_eccv,li2021hsmaldetailedhorseshape,bolanos_three-dimensional_2021,Zuffi_2024_CVPR}.
These representations, typically derived from 3D scans, are steered by body-joint rotation parameters and a latent shape code.

However, extracting the parameters of such representations from images is challenging due to the ill-posed nature of 3D inference from 2D images, the diversity of real-world animals, and the variability of imaging conditions in natural environments.
Training an effective regression model typically requires large amounts of annotated data.
Datasets curated for human pose and shape estimation, for parametric body models like SMPL~\cite{loper2015smpl}, include extensive sequences captured using marker-based mocap systems~\cite{h36m_pami} or fitted IMU devices~\cite{Marcard_2018_ECCV}.
While humans can be brought into 4D capture halls and outfitted with dense markers, wild animals are less cooperative.
As a result, alternative approaches must be devised to overcome the problem of obtaining data.

The most common approach is the manual annotation of anatomical 2D landmarks or silhouettes~\cite{biggs_2019}, which can be used for indirect model supervision.
Other approaches begin with 2D annotations and produce pseudo-labels of pose and shape by optimizing a model such as SMAL~\cite{Zuffi_2017_CVPR} to fit the labeled 2D features~\cite{xu2023animal3d}.
However, while optimizing SMAL to conform to a silhouette may yield 3D fits that appear plausibly aligned, producing accurate 3D annotations is difficult without sufficiently strong priors, as many physically implausible pose and shape combinations could explain the same silhouette.

Recently, the use of video-game engines to produce rendered synthetic data has been explored as an alternative to labeling real-world images~\cite{black2023bedlam, hewitt2023procedural, hu2019sail}.
While the approach sacrifices visual realism, it offers greater control over dataset curation.
Graphics engines' explicit representation of 3D scenes enables the production of precise ground-truth annotations and control over dataset statistics and variability.
However, traditional rendering-based synthetic datasets require substantial manual effort to design or modify.
To produce synthetic data for additional species, or render them in a new environment, requires a new set of 3D assets.
While synthetic data can be made to appear realistic, achieving both visual realism and sufficient diversity requires considerable resources.

We investigate a potentially simpler alternative to the production of synthetic data: \textit{rendering} with a conditional image-generation model.
We propose a pipeline that, given the name of a species, produces paired images and ground-truth pose-and-shape parameters.
Rather than relying on explicit collections of 3D assets and scenes, our pipeline is controlled by language: the inclusion of a new species or environmental setting is accomplished via prompting.
Our pipeline facilitates the generation of realistic images with a degree of control comparable to traditional synthetic-data generators, thus combining the advantages of visual realism, scalability, and controllable data production.

To demonstrate the scalability of our approach, we introduce \methodtitleshort, a million-image dataset comprised of unique poses and shapes across diverse mammalian quadrupeds; see \cref{fig:teaser} for examples.
Training a regression model solely on our synthetic dataset, without the use of ground-truth real-world images, we achieve state-of-the-art performance on Animal3D~\cite{xu2023animal3d}, a real-world animal pose and shape estimation benchmark, validating the quality of our dataset.
We also introduce a new synthetic evaluation dataset with greater annotation fidelity than previous benchmarks.

\noindent In summary, our key contributions include:
\begin{enumerate}
    \item A scalable pipeline for generating synthetic 3D animal pose and shape estimation data
    \item \methodtitleshort, a million-scale dataset of samples produced using our pipeline
    \item A state-of-the-art generic animal pose and shape regression model
    \item \testdatasettitle, a high-quality synthetic test dataset.
\end{enumerate}

\begin{figure*}[t]
\centering
\includegraphics[width=\linewidth]{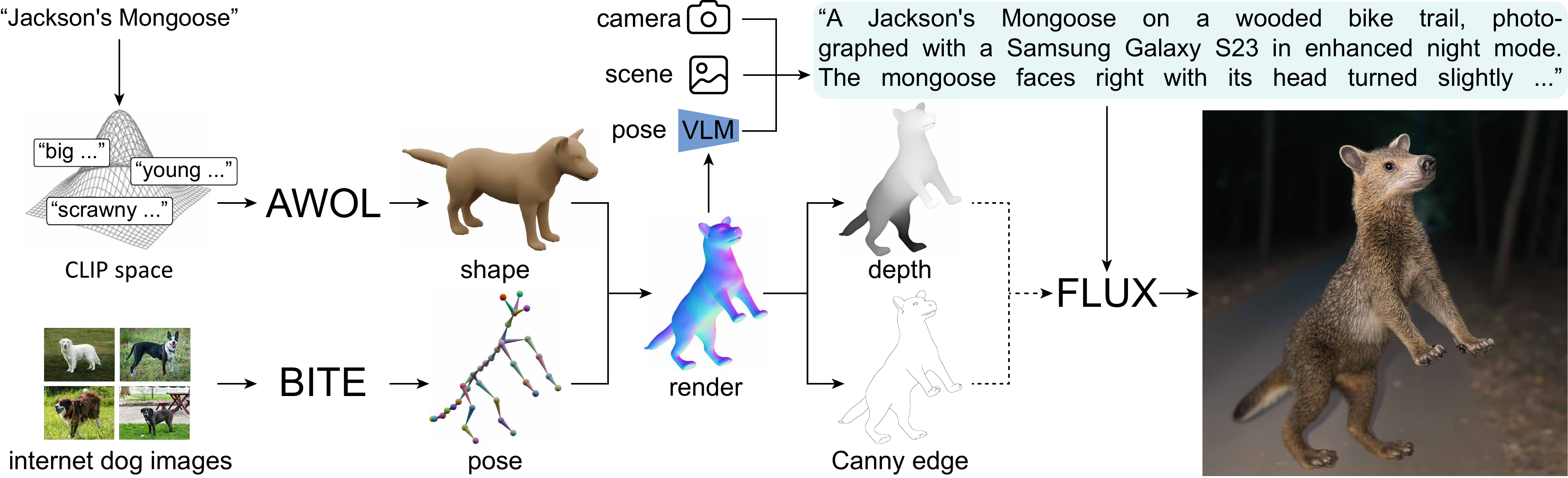}
\caption{
\textbf{Pipeline Overview.} Starting with a sampled animal name (\cref{ssec:species}), we sample corresponding shape parameters (\cref{ssec:shape}). Paired pose parameters are sampled from a set of pseudo-poses (\cref{ssec:pose}).
Sampled camera and scene descriptions are combined with a pose caption to form a prompt (\cref{ssec:prompt}).
Rendered control signals and the prompt are used to guide the conditional image-generation model, resulting in the final image (\cref{ssec:generation}).
}

\label{fig:pipeline}
\end{figure*}

\section{Related Work}\label{sec:related}
\smallskip\noindent\textbf{Animal Pose and Shape Estimation.}
The 3D reconstruction of animals follows two primary paradigms: model-free and model-based.
Model-free approaches make no assumptions about the animal's 3D body structure, and the objective is to obtain a representative 3D surface.
Given the diversity observed across animal species and shape, this approach is fairly common.
Notable examples include CMR~\cite{cmrKanazawa18}, which deformed a spherical mesh to reconstruct birds from images, and LASSIE~\cite{jampani_lassie} and MagicPony~\cite{Wu_2023_CVPR}, which learned articulated 3D shape from image collections.
ViSER~\cite{yang2021viser}, LASR~\cite{yang2021lasr}, BANMo~\cite{BANMO}, and PPR~\cite{yang2023ppr} recovered 3D shape and motion of animals from video.

Model-based approaches alternatively assume that a 3D model is provided (or retrievable~\cite{liu_casa}), either as a species-specific template model or as a parametric 3D model that captures shape variations within and across species. 
This approach is particularly valuable for downstream analysis, as 3D pose and shape parameters can be leveraged to estimate and track conformation and behavior over time. 
Among model-based approaches, \citet{fitzgibbon_dolphin_shape} were the first to address the modeling challenge, creating a morphable model for dolphins. 
Later, \citet{kanazawa_cats} estimated a deformable 3D model for cats. 
\citet{Zuffi_2017_CVPR} introduced SMAL, an articulated shape model for a variety of quadrupeds learned from scans of toy figures.
SMAL has been adopted to estimate shape and pose for zebras~\cite{Zuffi_2019_ICCV}, to estimate the 3D pose and shape of dogs~\cite{biggs_2019, biggs2020left,barc2022rueegg} from 2D datasets with keypoints and silhouette annotations, and to create dog avatars~\cite{sabathier_2024_eccv}.
Further modeling developments include \citet{ruegg2023bite}'s BITE extension of SMAL for dog breeds, \citet{li2021hsmaldetailedhorseshape}'s use of a horse-specific model for lameness detection, \citet{Wang_2021_CVPR}'s learning of a bird model from images, and \citet{Zuffi_2024_CVPR}'s pioneering horse model learned from real 4D scans.
\citet{zuffi2024awol} in AWOL introduced SMAL+, an enhanced version of the SMAL model learned from additional 3D scans.
\citet{kulits2024raw}'s RAW extended the animal-reconstruction problem to additionally model the surrounding environment.

\smallskip\noindent\textbf{Rendered Synthetic Data for Pose Estimation.}
Several works approach the generation of synthetic data for human pose and shape estimation.
SURREAL~\cite{varol17_surreal} sampled random backgrounds and applied cloth textures to posed SMPL~\cite{loper2015smpl} meshes.
AGORA~\cite{patel2021agora} rendered images of clothed-body scans with SMPL-X~\cite{pavlakos2019expressive} ground-truth annotations.
BEDLAM~\cite{black2023bedlam} extended this, presenting a dataset with simulated clothing, hair, and large variation in human shape.
\citet{hewitt2023procedural, hewitt2024look} applied displacement maps on a modified SMPL body to simulate natural cloth wrinkles.

In contrast, synthetic animal datasets lack comparable sophistication.
This is due in part to the lack of extensive motion datasets like AMASS~\cite{AMASS:ICCV:2019} and the greater morphological variation between animals.
Several methods have rendered a single species from a mesh to train a 2D joint regressor, including for mice, cougars, and dogs~\cite{bolanos_three-dimensional_2021, fangbemi2020zoobuilder, shooter2024sydog}.
However, none of these methods can be easily combined to train a multi-animal 3D pose and shape regressor.
Although \citet{mu2020learning} produced a dataset of more than ten different animals, it can only be applied to learn 2D joint estimation and not 3D pose and shape, as it lacks variation in shape.

\smallskip\noindent\textbf{Generative Models and Training Data.}
Recent advances in controllable image generation, such as Stable Diffusion~\cite{rombach2022high} and ControlNet~\cite{zhang2023adding}, enable realistic data generation for downstream tasks~\cite{azizi2023synthetic,ma2024generating,weng2024diffusion}.
DatasetDM~\cite{wu2023datasetdm} finetuned the Stable Diffusion decoder to generate synthetic images and ground-truth pairs for depth, human pose estimation, and semantic and instance segmentation.
Other works have synthesized data for human pose and shape estimation, generating novel images \cite{weng2024diffusion,zhang2023adding,ge2024d} or refining pre-rendered ones~\cite{cuevasvelasquez2024Human}.
While the use of generative models have been applied to the training of human pose and shape regression models, no prior work has explored the use of generative models for 3D animal pose and shape estimation.

\section{Method}\label{sec:method}

\begin{figure*}[t]
\centering
\rotatebox{90}{\hspace{0.75cm} Ours \hspace{1.7cm} GT \hspace{1.65cm} Input}
\includegraphics[width=0.975\linewidth]{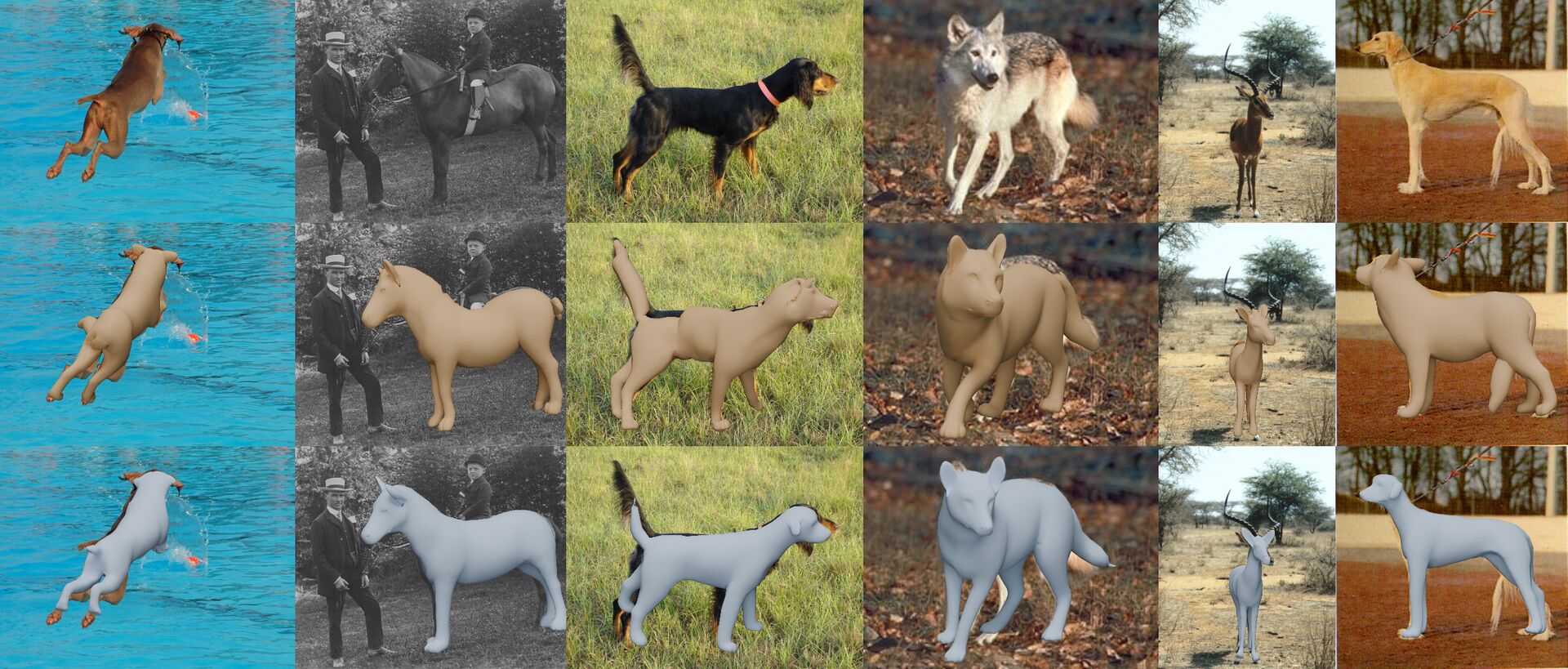}
\caption{\textbf{Animal3D Reconstruction Samples.} We show the input image (top), GT mesh (middle), and our model's prediction (bottom).}
\label{fig:reconstructions_animal3d-felidae}
\end{figure*}

In this section, we present our approach for generating synthetic training data for 3D animal pose and shape estimation (see \cref{fig:pipeline}).
After introducing the SMAL body model (\cref{ssec:smal}), which defines our pose and shape representation, we introduce our pipeline.
Starting with a set of mammalian species or breeds (\cref{ssec:species}), we sample a taxon.
Based on the taxon, we sample an animal shape (\cref{ssec:shape}) and assign a pose (\cref{ssec:pose}).
From the model parameters, we render a primitive image, which is captioned using a vision--language model (VLM) and used to synthesize a prompt (\cref{ssec:prompt}).
Finally, we condition an image-generation model using both the prompt and render (\cref{ssec:generation}).
We close with an explanation of our regression-model baseline (\cref{ssec:regressor}).

\subsection{SMAL}\label{ssec:smal}
The \textit{Skinned Multi-Animal Linear} (SMAL)~\cite{Zuffi_2017_CVPR} model is a function that, given shape parameters $\beta$ and pose parameters $\theta$, transforms a 3D template to produce a posed mesh.
The transformation occurs in two steps: first, the vertex template instance $\textbf{v}_{t}$ is deformed into an intrinsic shape $\textbf{v}_{s}$, then Linear Blend Skinning (LBS) is applied to rotate the deformed body parts based on the pose parameters $\theta$:
\begin{eqnarray}
\textbf{v}_{s} & = & \textbf{v}_{t} + B \beta^T \nonumber \\
\textbf{v} & = & \text{LBS}(\textbf{v}_{s}, \theta; W, J_r). 
\label{eq:model}
\end{eqnarray}

Here, the template $\textbf{v}_{t}$ represents the initial state of a triangular mesh with $n_V$ vertices, $B$ is a matrix of shape $3n_V{\times}n_B$ containing the $n_B$ basis vectors of a linear shape deformation space, $J_r$ is the joint regressor that maps model vertices to a set of $n_J$ 3D joint locations, and $W$ is a skinning weight matrix used in LBS.

The linear shape space is learned using Principal Component Analysis (PCA) on a set of scans of toy quadrupedal animals.
In particular, we employ SMAL+, an expanded variant of SMAL introduced in AWOL~\cite{zuffi2024awol}, learned from a set of $145$ registered scans.

\subsection{Species Sampling}\label{ssec:species}
Our approach offers a key advantage over traditional rendering-based synthetic-data generation: rather than requiring additional artist-designed 3D assets, adding a new species requires only prompt modifications.
This enables precise control over dataset sampling statistics: taxon can be readily added or proportions re-balanced.

\begin{figure}[t]
\centering 
\includegraphics[width=\linewidth]{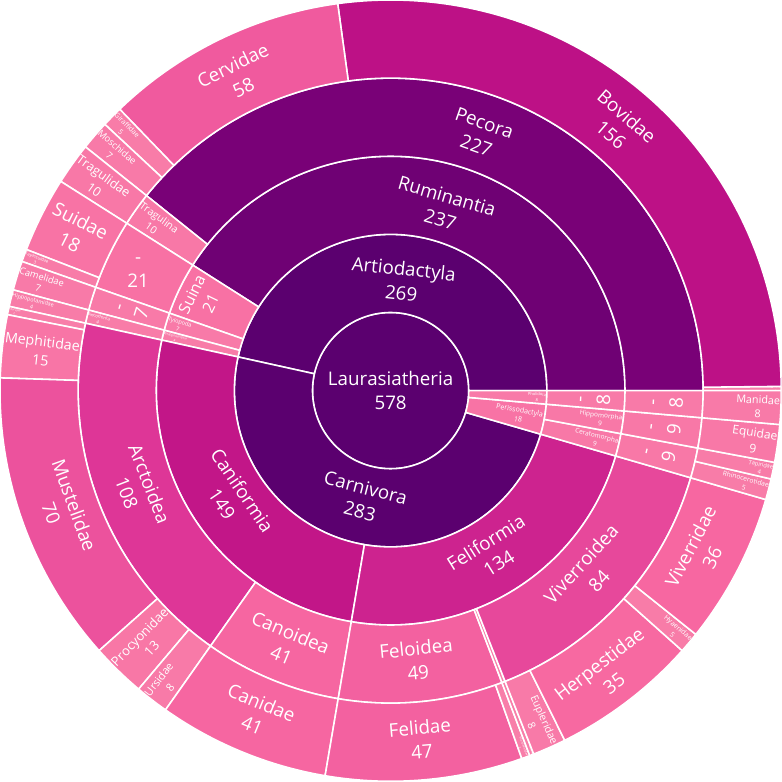} 
\caption{ \textbf{Taxonomy.} We sample species from a subset of the mammalian Superclass Laurasiatheria. The figure displays the abbreviated taxonomical hierarchy of our sampling, where hyphens represent an empty level and the numbers are of contained species.
} 
\label{fig:taxonomy} 
\end{figure}

To maximize data diversity, we sample from a variety of taxa listed in the Mammal Diversity Database~\cite{mammal_diversity_database}.
While SMAL can represent a wide range of quadrupedal animals, its fixed joint topology and limitated shape-space expressivity constrain the set of representable taxa.
As a result, we restrict our sampling to mammals within the superorder Laurasiatheria, excluding members of the order Eulipotyphla. \cref{fig:taxonomy} shows the resulting taxonomy.

We note that breeds are recognized distinctly from taxonomical species.
For example, \textit{Canis familiaris} (dog) is regarded as a single species.
Considering the diversity in shape, size, and visual appearance across breeds of dogs, we opt to combine species sampling with a set of 247 dog breeds.
We implement a balanced sampling strategy where dog breeds collectively have equal weight to species, with a 50\% probability of any given sample being a dog breed.

\subsection{Shape Sampling}\label{ssec:shape}
AWOL~\cite{zuffi2024awol} is a recent flow-based generative model that maps CLIP~\cite{radford2021learning} embeddings to SMAL shape parameters (betas), generating shapes conditioned on an embedded text prompt of an animal name or description.
While AWOL often generates shapes that are well-aligned with a given prompt, it is non-stochastic, and each embedding maps to only one shape.
This presents a challenge for well-covering the space of possible shapes.

One might consider avoiding AWOL altogether and sampling betas naively.
However, this can result in implausible shapes and limits data realism.
In this case, shape realism and diversity are competing objectives.

Instead, we opt to sample in CLIP embedding space and use AWOL for decoding, striking a balance between realism and diversity.
For each taxon, we compute 128 CLIP text embeddings, using a list of appearance descriptors such as ``big,'' ``young,'' or ``scrawny.''
CLIP is prompted with ``A photo of a X Y.''
We fit a multivariate Gaussian distribution to the resulting embeddings from which we sample to generate stochastic shapes that maintain alignment with the taxon.
This preserves shape plausibility while introducing controlled variability.
See \cref{fig:pipeline} for a visualization.

\begin{table}[t]
\centering
{
\setlength{\tabcolsep}{2pt}
\begin{tabular}{lccc}
\toprule
 & $\uparrow$ PCK@0.5 & $\downarrow$ S-MPJPE & $\downarrow$ PA-MPJPE \\
\midrule
Ours & \textbf{97.0} & \textbf{160.1} & \textbf{116.6} \\
Ours (ResNet) & \underline{95.11} & \underline{201.1} & 132.67 \\
\midrule
HMR$^*$ & 63.1 & 496.2 & 124.8 \\
PARE$^*$ & 85.6 & 374.9 & 127.2 \\
WLDO$^*$ & 65.1 & 484.0 & \underline{123.9} \\
\bottomrule
\end{tabular}
}
\caption{\textbf{Quantitative Method Comparison}. We compare our models trained on \methodtitleshort with the best-performing numbers on the Animal3D benchmark.
Asterisks ($*$) signify the result is from Animal3D and the best number was chosen across experiments.
}
\label{table:evaluation}
\end{table}

\subsection{Pose Sampling}\label{ssec:pose}
A similar dilemma between diversity and realism arises in the sampling of ground-truth pose.
To maximize diversity, one could sample random rotations.
This is feasible with human body models like SMPL~\cite{loper2015smpl} where vast amounts of motion-capture data are available, which can be used to learn a generative pose prior~\cite{pavlakos2019expressive}.
However, the absence of comparable data for animals or SMAL makes sampling valid poses more difficult and requires different strategies.

To address this data limitation, we apply BITE~\cite{ruegg2023bite}, an optimization-based dog-pose estimation method, to process an expansive collection of online dog images and extract plausible pseudo-poses.
Although BITE does not extract accurate poses from every image, we use only the extracted pose, discarding the original photos.
We then sample from this collection of poses for our dataset generation.
While the same set of rotations may produce different resulting poses on animals of varying proportions, we find that dog poses generally transfer reasonably across different quadrupeds.

\subsection{Prompt Sampling}\label{ssec:prompt}
After sampling species, shape, and pose, we render out the resulting model in Pyrender~\cite{pyrender}.
The render is passed to a VLM, Molmo-7B-D-0924~\cite{deitke2024molmopixmoopenweights}, prompted with ``This is a picture of an animal. Which direction is the animal facing?''
The caption provides guidance on global orientation, helping to reduce ambiguity in the subsequent image-generation step, particularly for complex poses.

In addition to captioning, and inclusion of the species name, we also sample a camera setting (e.g., ``Samsung Galaxy S23 with enhanced night mode camera'') and a scenery setting (e.g., ``bike trail through the woods'') from predefined lists.
We find that explicit prompting of these attributes increases visual realism and diversity.
After the above are sampled, we pass the descriptors to an LLM, Qwen2.5-7B-Instruct~\cite{qwen2_5}, to synthesize these components into a coherent and concise prompt.
See \cref{ssec:prompt_ablation} for a quantitative ablation study on our prompting design decisions.

\begin{figure*}[t]
\raggedright
\hspace{0.6cm} Input \hspace{1.45cm} GT \hspace{0.8cm} HMR-Human$^*$ \hspace{0.1cm} HMR-Syn$^*$ \hspace{0.1cm} PARE-Human$^*$ \hspace{0.05cm} PARE-Syn$^*$ \hspace{0.2cm} WLDO-Syn$^*$ \hspace{0.7cm} Ours
\includegraphics[width=\linewidth]{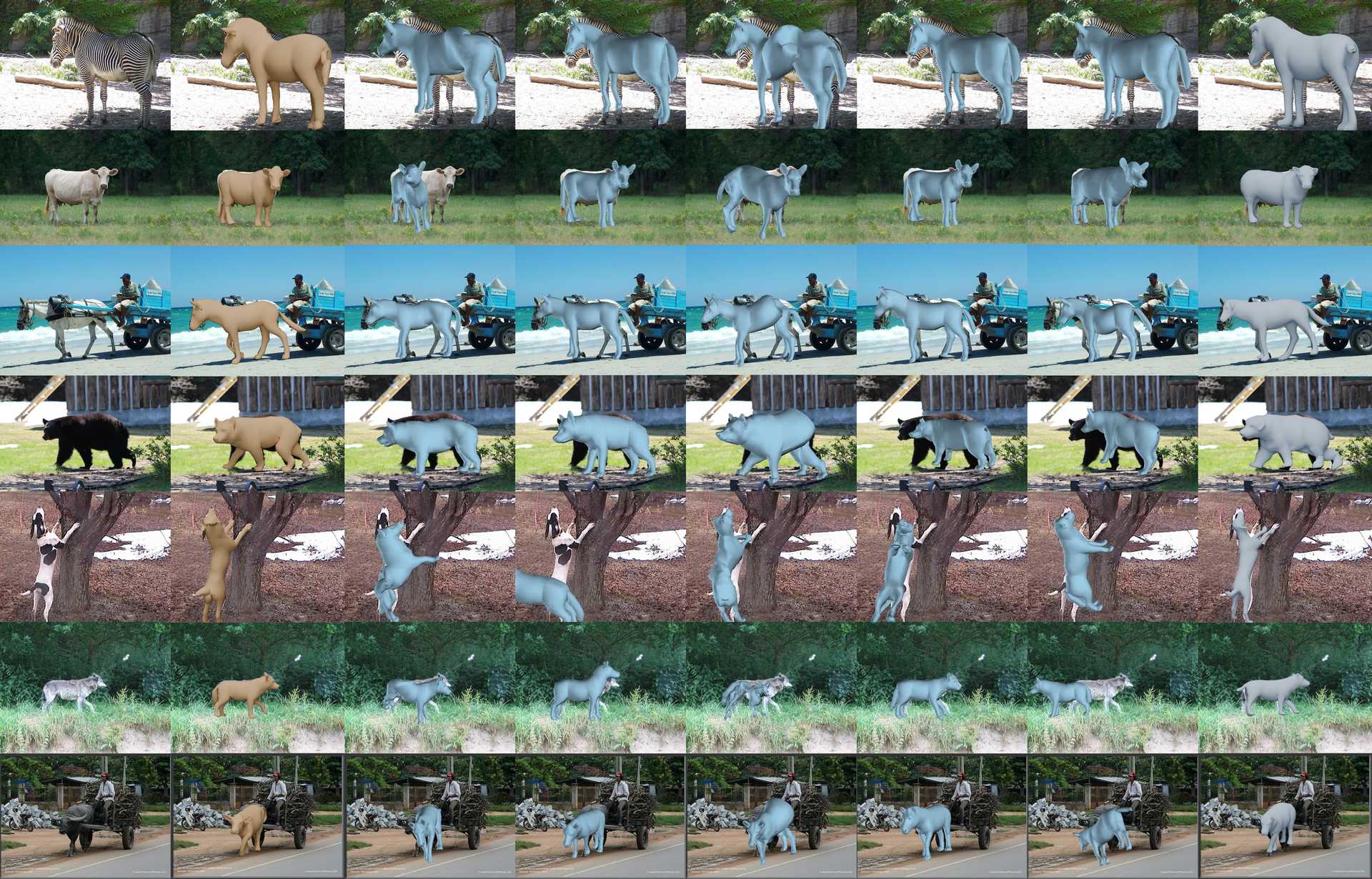}
\caption{
\textbf{Qualitative Method Comparison.} Predictions between our method and the baseline results (*) sourced from Animal3D~\cite{xu2023animal3d}.
}
\label{fig:comparison}
\end{figure*}

\subsection{Conditional Image Generation}\label{ssec:generation}
Once the prompt has been prepared, we employ FLUX~\cite{flux}, a generative text-to-image diffusion model, to synthesize paired images.
While FLUX can be conditioned on CLIP~\cite{radford2021learning} and T5~\cite{t5} embeddings, it cannot natively be controlled with pose and shape parameters.
To achieve such control, we employ an auxiliary ControlNet~\cite{zhang2023adding} model.

ControlNets are fine-tuned versions of a base image-diffusion model -- FLUX in our case -- trained to maximize output probability conditioned on provided control signals, such as a Canny-edge or depth map.
Using Pyrender, we produce a depth map and a shaded render for Canny-edge extraction.
During generation, FLUX and the ControlNet are used concurrently to guide the diffusion process by both the text prompt and extracted control signals.
We generate all images at a resolution of 1024x1024.
See \cref{fig:pipeline} for a visual of the control signals applied, \cref{ssec:control_ablation} for an ablation on the control signal used, and also Supp.\ Mat.\ for an ablation on the choice of image-generation model.

\subsection{Parameter Regressor}\label{ssec:regressor}
We train a regression model on \methodtitleshort using two architectures.
Following the results of an ablation study performed by \citet{Goel_2023_ICCV}, we adopt ViTPose~\cite{NEURIPS2022_fbb10d31} as the primary backbone of our regressor.
This backbone was originally trained for human 2D-keypoint estimation.
To match the baselines employed in Animal3D~\cite{xu2023animal3d}, we additionally train our model with ResNet-50~\cite{He_2016_CVPR}.
We supervise the training of the model using losses on 2D-joint projection, pose parameters, and shape.
See Supp.\ Mat.\ for additional details.
\begin{table*}[b]
\centering
{
\begin{tabular}{lccc|cc}
\toprule
& \multicolumn{3}{c}{Animal3D} & \multicolumn{2}{c}{GenZoo-Felidae} \\
\cmidrule(lr){2-4} \cmidrule(lr){5-6}
 & $\uparrow$ PCK@0.5 & $\downarrow$ S-MPJPE & $\downarrow$ PA-MPJPE & $\downarrow$ S-V2V & $\downarrow$ PA-V2V \\
\midrule
Full & \underline{97.1} & \textbf{166.9} & \textbf{118.4} & \underline{59.3} & \underline{50.2} \\
\midrule
-Depth & 96.7 & 184.1 & 135.1 & 95.4 & 65.9 \\
-Canny & 96.2 & 172.3 & \underline{119.4} & \textbf{57.7} & \textbf{39.1} \\
-Caption & 96.9 & \underline{167.1} & 120.1 & 71.0 & 48.6 \\
-LLM & \textbf{97.2} & 168.2 & 120.7 & 69.4 & 49.7 \\
\bottomrule
\end{tabular}
}
\caption{\textbf{Quantitative Ablation Effects.} Ablation-study results for models trained on 100,000 samples each.
}
\label{table:ablation}
\end{table*}

\section{Evaluations}\label{sec:evaluations}
We evaluate our method quantitatively using three model-joint metrics: 1) PCK@0.5, defined as the percentage of correct keypoints within half the head--tail length of the ground truth; 2) PA-MPJPE, the procrustes-aligned mean per-joint positional error; and 3) S-MPJPE, defined as PA-MPJPE without the rotation transform, as used in Animal3D~\cite{xu2023animal3d} to account for variations in SMAL scale.

\subsection{Animal3D}\label{ssec:animal3d}
We employ Animal3D~\cite{xu2023animal3d} to evaluate the transferability of our model trained on \methodtitleshort.
Animal3D is built on a set of images borrowed from the ImageNet~\cite{imagenet} and COCO~\cite{coco} datasets. 
Annotators manually labeled 2D keypoints and silhouettes, which were used to guide an optimization-based fitting process, resulting in SMAL pseudo-labels.

Training solely on \methodtitleshort, we achieve state-of-the-art performance, outperforming the best baseline by 57\% in S-MPJPE (374.9$\rightarrow$160.1).
We also observe notable improvements in PCK@0.5 (85.6$\rightarrow$97.0) and PA-MPJPE (123.9$\rightarrow$116.6), but gains appear comparatively saturated. 
PCK@0.5, which we report in order to compare with the numbers reported in Animal3D, is a very generous metric.
Despite large improvement seen in S-MPJPE, we find only modest improvement over the baselines in PA-MPJPE.

\begin{figure}[t]
\centering
\hspace{-0.5cm} Input \hspace{1.9cm} GT \hspace{2.25cm} Ours
\includegraphics[width=\linewidth]{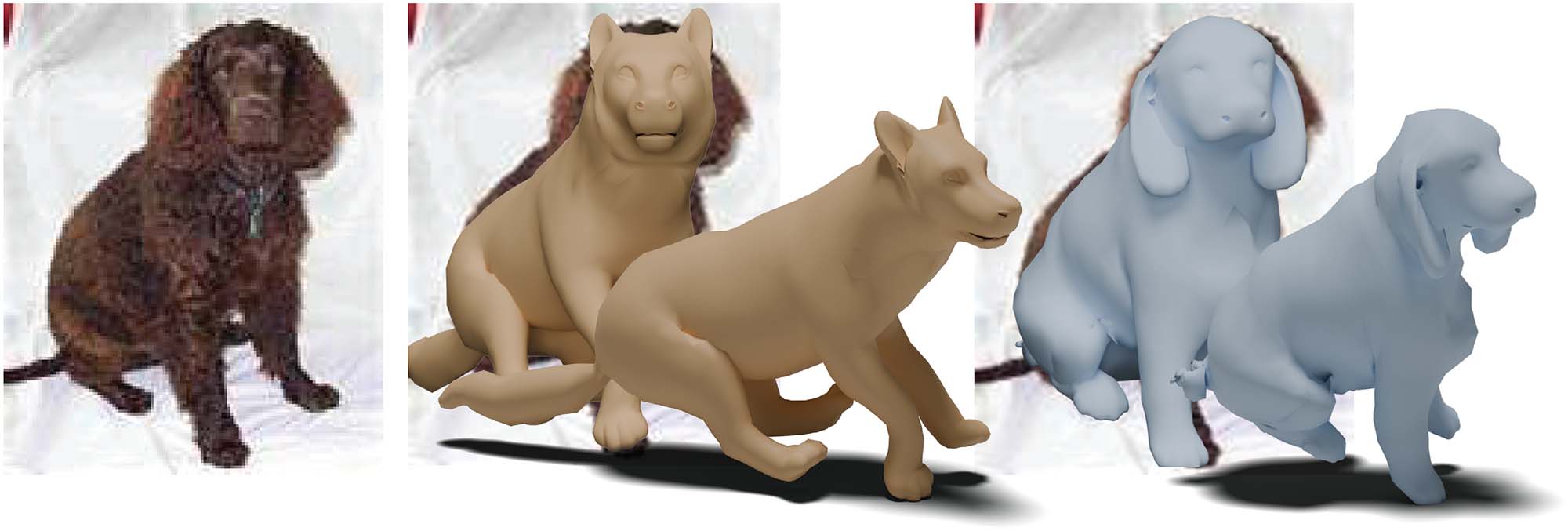}
\caption{
\textbf{Animal3D Comparison.} Highlighting the difficulty of producing manual 3D annotations of monocular, real-world images, we observe physical implausibilities in the Animal3D ground-truth.
In contrast, a model trained on our dataset does not learn the same biases.
See Supp.\ Mat.\ for a perceptual study comparing our method and the dataset's manual annotations.
}
\label{fig:perceptual_motivation}
\end{figure}

To investigate this, we look to the ground-truth data.
We visualize in \cref{fig:perceptual_motivation} a ground-truth Animal3D annotation next to our model prediction.
While we observe that the projection of the ground-truth mesh is well-aligned with the image silhouette, when viewed from the side it appears to be a very different animal, highlighting the difficulty of producing accurate manual annotations without sufficiently strong priors.
This suggests that there may be an upper bound on Animal3D performance.
We explore this further through a perceptual study detailed in the  Supp.\ Mat.

\subsection{GenZoo-Felidae}\label{ssec:genzoo-felidae}

\begin{figure}[t]
\centering
\begin{tabular}{@{}c@{}c@{}c@{}c@{}}
\makebox[0.25\linewidth][c]{\small Render} &
\makebox[0.25\linewidth][c]{\small Canny} &
\makebox[0.25\linewidth][c]{\small Canny+Depth} &
\makebox[0.25\linewidth][c]{\small Depth}\\
\includegraphics[width=0.25\linewidth]{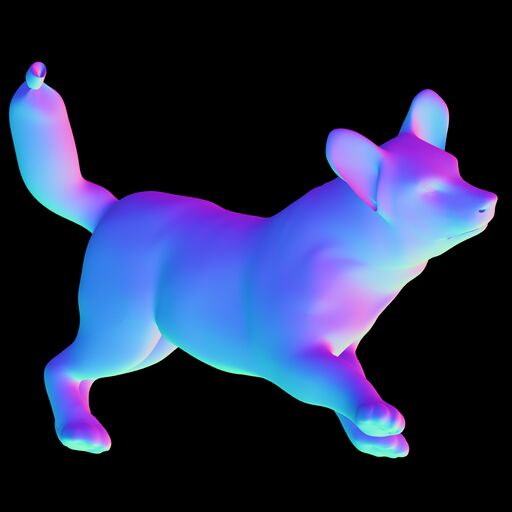} &
\includegraphics[width=0.25\linewidth]{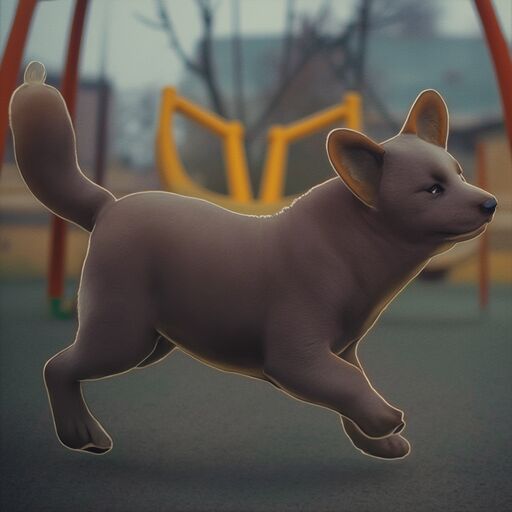} &
\includegraphics[width=0.25\linewidth]{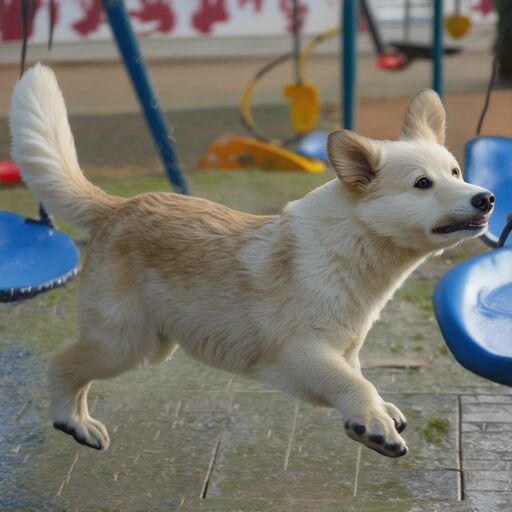} &
\includegraphics[width=0.25\linewidth]{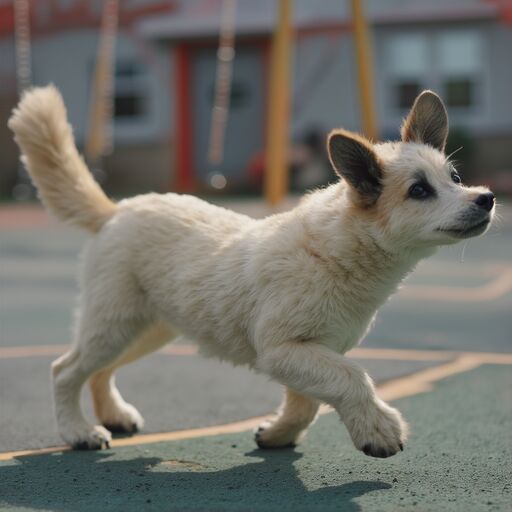}
\end{tabular}
\caption{\textbf{Control-Signal Ablation.}
Depth-only conditioning produces the most realistic images but poorest alignment with ground-truth poses. 
Canny-edge-only control shows the opposite effect.
We employ both to balance visual realism with pose accuracy.
}
\label{fig:control_vis}
\end{figure}

\begin{figure}[t]
\centering
\includegraphics[width=\linewidth]{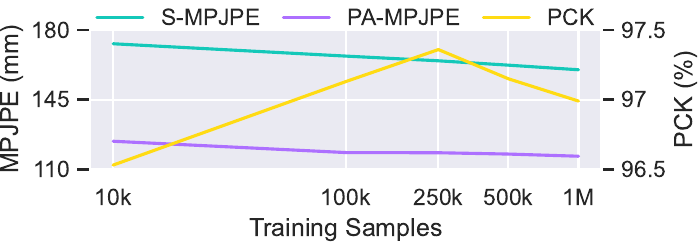}
\caption{
\textbf{Data Efficiency.} We evaluate data efficiency of our model and observe consistent log-linear trends as data is scaled.
}
\label{fig:data_efficiency}
\end{figure}

\begin{figure*}[t]
\centering
\rotatebox{90}{\hspace{1.05cm} Ours \hspace{2.1cm} GT \hspace{2.0cm} Input}
\includegraphics[width=0.975\linewidth]{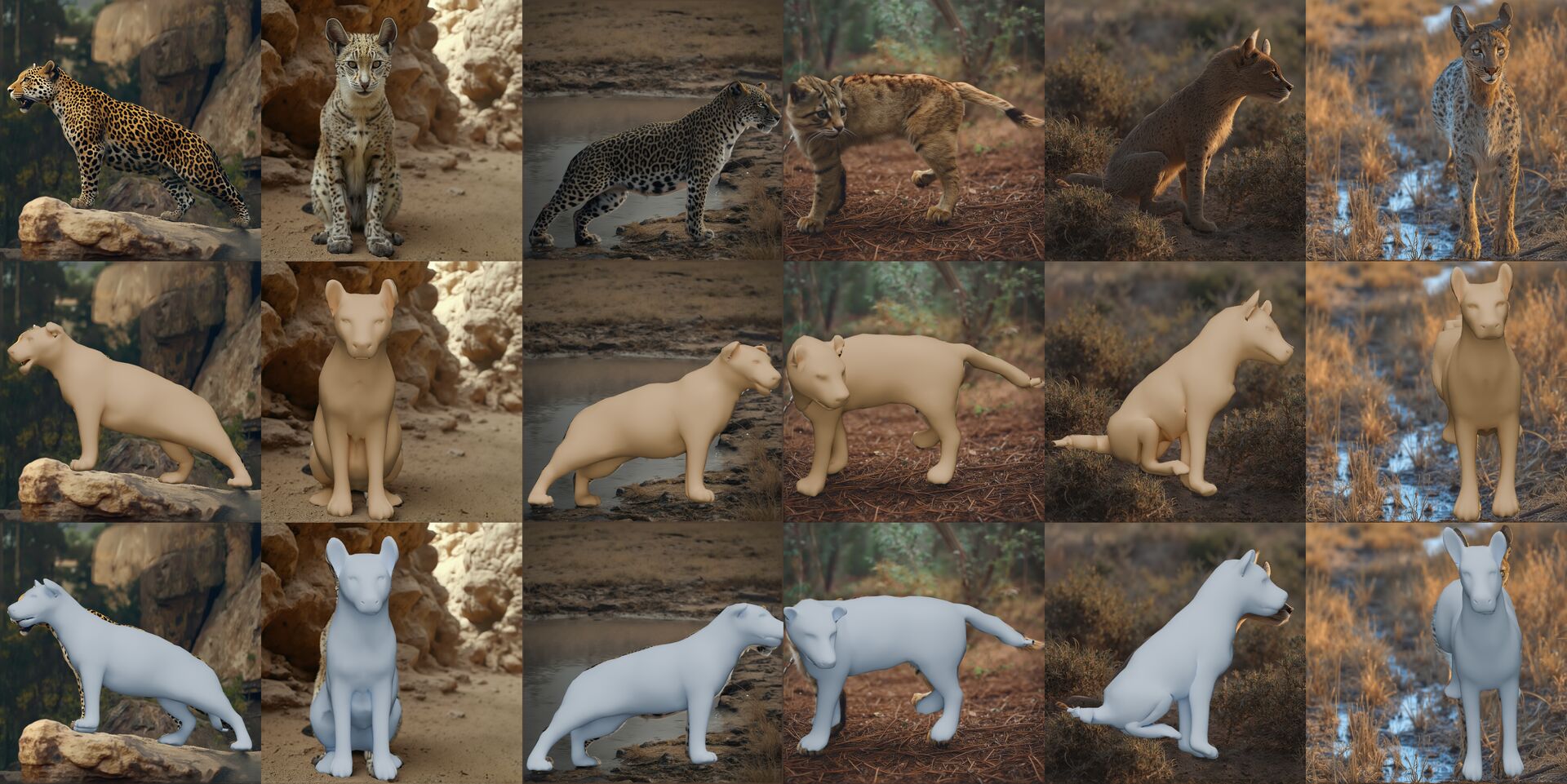}
\caption{\textbf{GenZoo-Felidae Dataset Samples.}
We show the input image (top), GT mesh (middle), and our model's prediction (bottom).
}
\label{fig:reconstructions_genzoo-felidae}
\end{figure*}

Motivated by our observations in \cref{ssec:animal3d}, we propose a complementary evaluation dataset.
We design it both to benchmark model generalization to unseen species and to evaluate shape-estimation performance against ground-truth shape.
We split \methodtitleshort to create a 100-sample test set with no overlap in species, pose, camera setting, or environmental conditions.
We designate the 47 species in the family Felidae as test-only and do not train on them in any evaluation of the paper.
We refer to our benchmark as \testdatasettitle.
We show dataset samples and model reconstructions in \cref{fig:reconstructions_genzoo-felidae}, and we report metrics in \cref{table:ablation}.

\subsection{Control Signal}\label{ssec:control_ablation}
We evaluate different ControlNet conditioning combinations, comparing our full model (combined Canny-edge and depth control) against the controls applied individually.
Depth-only conditioning produces the most-realistic images but shows poorest ground-truth alignment.
In contrast, Canny-edge-only conditioning achieves the best alignment but compromises visual realism.
We observe that combining the two at reduced strength balances these trade-offs. 
See \cref{fig:control_vis} for an illustrative sample.
We find that quantitative evaluation supports our qualitative findings (\cref{table:ablation}).

\subsection{Captioning and Textual Conditioning}\label{ssec:prompt_ablation}
We assess the impact of removing the VLM captioning step and LLM prompt synthesis.
While both components positively contribute to Animal3D performance metrics, their impact is most pronounced in \testdatasettitle.
See \cref{table:ablation}.

\subsection{Data Efficiency}\label{ssec:data_efficiency}
We evaluate our model trained with varying amounts of data.
While we observe a positive quantitative trend as the amount of training data increases, the returns are diminishing.
This further suggests that there may be an upper bound on Animal3D performance.
See \cref{fig:data_efficiency} for a log-linear plot.

\section{Discussion and Limitations}\label{sec:limitations}
While our \methodtitleshort-trained model demonstrates fairly robust generalization to real-world images, several limitations warrant discussion:
1) Our model struggles under strong occlusions, such as mistaking a human in the foreground as the regression target.
Future work should explore methods for synthesizing realistic occlusions in training data to improve robustness.
See Supp.\ Mat.\ for a 100+ page comprehensive set of Animal3D visualizations, including method failures.
2) While our pose-sampling distribution, built from pseudo dog poses, appears sufficient for a broad coverage of animal poses, the model struggles with species-specific poses not typically observed in dogs, such as feline grooming positions.
Future work should explore more-comprehensive pose-sampling strategies.
3) Although the SMAL model is capable of representing well a broad swathe of shapes and species, it is constrained by its fixed skeletal topology and cannot represent large morphological differences between species such as the trunk of an elephant.
Future work should focus on developing more-expressive parametric representations that can accommodate greater anatomical diversity.
4) We observe that FLUX has limited understanding relating to lesser-known species, often instead producing a more common, visually similar animal, but sometimes mistaking the name for a tropical bird.
Future work should explore the adaptation of image-generation models to better represent rare and unusual species.
5) While there is a lack of training data for 3D animal pose-and-shape estimation, there is also a need for strong benchmarks with precise annotations.

\section{Conclusion}\label{sec:conclusion}
\vspace{0.15cm}
Motivated by the shortcomings of existing approaches for the acquisition of 3D animal pose and shape estimation training data, we proposed a scalable pipeline that leverages conditional image-generation models.
Our pipeline enables the generation of realistic images with a degree of control comparable to that of traditional synthetic-data generators.
Showcasing the scalability of our approach, we presented \methodtitleshort, a dataset of one million images of unique animals.
Training solely on \methodtitleshort, without the use of any real-world training data, we demonstrated state-of-the-art performance on a real-world animal pose and shape estimation benchmark.
We additionally introduced \testdatasettitle, a high-quality synthetic test dataset that complements existing pseudo-labeled real-world evaluations.
Beyond immediate technical achievements, our work opens new possibilities for automated animal behavior analysis, wildlife monitoring, and veterinary applications.

{\small\noindent\textbf{Acknowledgments} We thank Tsvetelina Alexiadis and Taylor McConnell for study counsel.
\textbf{Disc.\ }MJB has received research gift funds from Adobe, Intel, Nvidia, Meta/Facebook, and Amazon.  MJB has financial interests in Amazon and Meshcapade GmbH.  While MJB is a co-founder and Chief Scientist at Meshcapade, his research in this project was performed solely at, and funded solely by, the Max Planck Society. SZ is supported by PNRR FAIR Future AI Research (PE00000013), Spoke 8 Pervasive AI (CUP H97G22000210007) and NBFC National Biodiversity Future Center (CN00000033), Spoke 4 (CUP B83C22002930006) under the NRRP MUR program by NextGenerationEU.}

\clearpage
\setcounter{page}{1}
\maketitlesupplementary
\appendix

\section{Training Details}\label{sec:hyperparameters}
Our model is trained with three losses: a joint 2D-projection L1 loss with weight \mbox{\texttt{0.01}}, an 9D-rotation-matrix MSE after performing symmetric orthogonalization with weights of \mbox{\texttt{100}} on \mbox{\texttt{body\_pose}} and \mbox{\texttt{global\_orient}} (following ablations performed in \cite{geist2024learning}), and an L1 loss on transformed vertices after applying \mbox{\texttt{betas}} with a weight of \mbox{\texttt{50}}.
A batch size of \mbox{\texttt{128}} with a single GPU is used across experiments.
We configure early stopping based on validation joint 2D-projection loss.

\section{Perceptual Study}\label{sec:perceptual}
Motivated by inconsistencies observed during evaluation on Animal3D~\cite{xu2023animal3d} (see Sec.\ 4.1 and Fig.\ 6), we perform a perceptual study comparing our predictions against the Animal3D ground truth to investigate our hypothesis that there is an upper limit on achievable quantitative performance.
We show 48 participants on Amazon Mechanical Turk (AMT) a set of 22 randomly selected dataset samples along with five warm-up samples and three catch trials.
Each sample consists of the source image and side renders of both the ground-truth and predicted meshes.
Participants are tasked with determining which of the two meshes is posed in a way that is better aligned with the animal in the image.
Warm-up samples are discarded prior to analysis, and the five participants that failed two or more catch trials are excluded.
However, we note that quantities reported below do not change when participants are not excluded.

For each of the samples, we perform a one-sided binomial test to determine whether the predicted mesh is preferred over the ground truth mesh.
We find that the predicted mesh is \textit{significantly} preferred at $\alpha = 0.05$ in 27\% of the samples.
To correct for multiple comparisons, we apply the Benjamini--Hochberg correction across tests, and find that the result remains significant.
In response, we reject the null hypothesis that ground-truth samples are consistently preferable.

\section{Image-Generation-Model Ablation}\label{sec:flux_ablation}
\begin{table*}[t]
\centering
{
\begin{tabular}{lccc|cc}
\toprule
& \multicolumn{3}{c}{Animal3D} & \multicolumn{2}{c}{GenZoo-Felidae} \\
\cmidrule(lr){2-4} \cmidrule(lr){5-6}
 & $\uparrow$ PCK@0.5 & $\downarrow$ S-MPJPE & $\downarrow$ PA-MPJPE & $\downarrow$ S-V2V & $\downarrow$ PA-V2V \\
\midrule
FLUX & \underline{97.1} & \textbf{166.9} & \textbf{118.4} & \textbf{59.3} & \underline{50.2} \\
Hunyuan-DiT & 95.9 & \underline{174.0} & \underline{125.6} & \underline{67.8} & \textbf{47.6} \\
Stable Diffusion 3 & \textbf{97.5} & 178.3 & 127.8 & 85.2 & 61.1 \\
\bottomrule
\end{tabular}
}
\caption{\textbf{Image-Generation-Model Ablation Effects}. We ablate our choice of image-generation model, generating datasets of 100k samples each.
We observe that FLUX outperforms the ablated models, Hunyuan-DiT and Stable Diffusion 3, in the majority of metrics.
}
\label{table:flux_ablation}
\end{table*}

\begin{figure}[t]
\centering
\begin{tabular}{@{}c@{}c@{}c@{}c@{}}
\makebox[0.25\linewidth][c]{\small Render} & 
\makebox[0.25\linewidth][c]{\small Flux} & 
\makebox[0.25\linewidth][c]{\small Hunyuan-DiT} & 
\makebox[0.25\linewidth][c]{\small SD3}\\
\includegraphics[width=0.25\linewidth]{figures/ablation/render_dog.jpg} &
\includegraphics[width=0.25\linewidth]{figures/ablation/cd_dog.jpg} &
\includegraphics[width=0.25\linewidth]{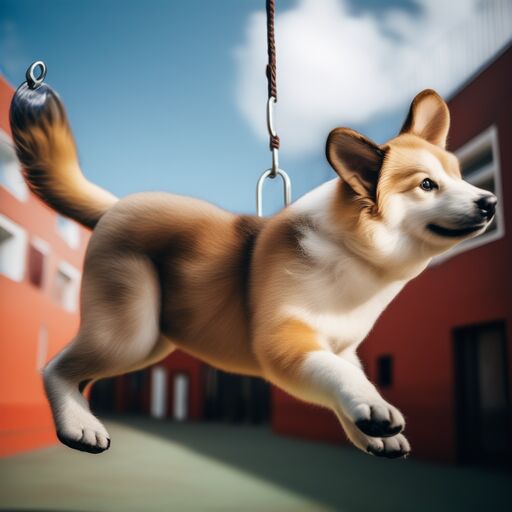} &
\includegraphics[width=0.25\linewidth]{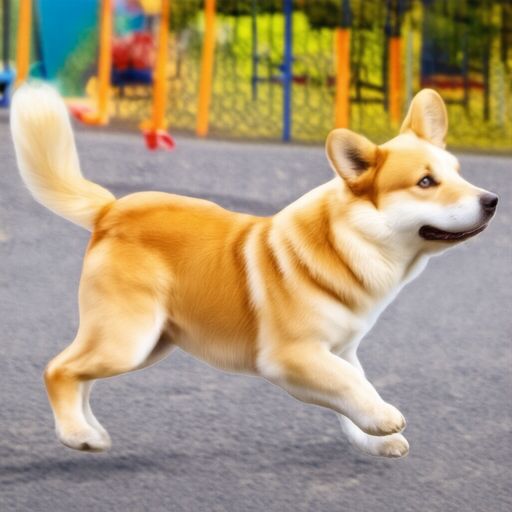}
\end{tabular}
\caption{\textbf{Image-Generation-Model Ablation.} We ablate our choice of FLUX~\cite{flux} as the image-generation model, comparing against Hunyuan-DiT~\cite{li2024hunyuanditpowerfulmultiresolutiondiffusion} and Stable Diffusion 3 (SD3)~\cite{esser2024scaling}.
We observe that both ablated models produce results that are less visually realistic than those of FLUX, and tend to additionally more-often exhibit unnatural artifacts.
Hunyuan-DiT appears to produce more cartoon-like samples, whereas we observe more-frequent control-signal failures with Stable Diffusion 3.
}
\label{fig:flux_ablation_vis}
\end{figure}

We additionally ablate our choice of FLUX as the image-generation model.
We compare against Hunyuan-DiT~\cite{li2024hunyuanditpowerfulmultiresolutiondiffusion} and Stable Diffusion 3~\cite{esser2024scaling}.
Training on 100,000 samples for each experiment, we observe greatest performance training on images produced using FLUX.
We report quantitative results in \cref{table:flux_ablation} and include a visual comparison in \cref{fig:flux_ablation_vis}.
While PCK scores remain somewhat saturated across experiments, as observed during earlier experiments, we observe more-notable differences in 3D metrics, with FLUX outperforming Hunyuan-DiT and Stable Diffusion 3 placing third.
We observe qualitatively that the Stable Diffusion 3 model can struggle producing outputs aligned with the control signal, and that both of the non-FLUX models produce comparatively unrealistic images.
Even without control signals, we find that Hunyuan-DiT produces samples that appear cartoon-like, while Stable Diffusion 3 generations are made more natural.

\section{Additional GenZoo Samples}
See \url{https://genzoo.is.tue.mpg.de} for additional dataset samples.

\section{Additional Animal3D Results}
See \url{https://genzoo.is.tue.mpg.de} for comprehensive Animal3D reconstructions.

{
    \small
    \bibliographystyle{ieeenat_fullname}
    \bibliography{main}
}

\end{document}